\documentclass{article}
\usepackage{PRIMEarxiv}
\usepackage{cite}
\usepackage{amsmath,amssymb,amsfonts}
\usepackage{lipsum}
\usepackage{graphicx} 
\newcommand{\tabincell}[2]{\begin{tabular}{@{}#1@{}}#2\end{tabular}}  
\usepackage{textcomp}
\usepackage{color}
\usepackage{xcolor}
\usepackage{array}
\usepackage{booktabs}
\usepackage{tabularx}

\usepackage{algorithm}  
\usepackage{algorithmicx}  
\usepackage{algpseudocode}

\def\BibTeX{{\rm B\kern-.05em{\sc i\kern-.025em b}\kern-.08em
    T\kern-.1667em\lower.7ex\hbox{E}\kern-.125emX}}

\usepackage{fancyhdr}

\begin{document}

\title{Knowledge-Informed Auto-Penetration Testing Based on Reinforcement Learning with Reward Machine
\thanks{This work is supported in part by the Natural Sciences and Engineering Research Council of Canada under grants RGPIN-2018-06724 and DGECR-00022-2018 and by the Concordia University Research Chair in Artificial Intelligence in Cyber Security and Resilience.
This work has been submitted to the IEEE World Congress on Computational Intelligence 2024. Copyright may be transferred without notice, after which this version may no longer be accessible.}}

\author{
  Yuanliang Li \\
  Concordia University \\
  Montr\'eal, Canada\\
  \texttt{yuanliang.li@concordia.ca} \\
   \And
  Hanzheng Dai \\
  Concordia University \\
  Montr\'eal, Canada\\
  \texttt{hanzheng.dai@mail.concordia.ca} \\
   \And
  Jun Yan \\
  Concordia University \\
  Montr\'eal, Canada\\
  \texttt{jun.yan@concordia.ca} \\
   \And
}

\maketitle

\thispagestyle{fancy} 
\lhead{} 
\chead{} 
\rhead{} 
\cfoot{} 
\renewcommand{\headrulewidth}{0pt} 
\renewcommand{\footrulewidth}{0pt}

\begin{abstract}
Automated penetration testing (AutoPT) based on reinforcement learning (RL) has proven its ability to improve the efficiency of vulnerability identification in information systems. However, RL-based PT encounters several challenges, including poor sampling efficiency, intricate reward specification, and limited interpretability. To address these issues, we propose a knowledge-informed AutoPT framework called DRLRM-PT, which leverages reward machines (RMs) to encode domain knowledge as guidelines for training a PT policy. In our study, we specifically focus on lateral movement as a PT case study and formulate it as a partially observable Markov decision process (POMDP) guided by RMs. We design two RMs based on the MITRE ATT\&CK knowledge base for lateral movement. To solve the POMDP and optimize the PT policy, we employ the deep Q-learning algorithm with RM (DQRM). The experimental results demonstrate that the DQRM agent exhibits higher training efficiency in PT compared to agents without knowledge embedding. Moreover, RMs encoding more detailed domain knowledge demonstrated better PT performance compared to RMs with simpler knowledge.
\end{abstract}

\keywords{Penetration testing \and Reinforcement learning \and Human knowledge integration \and Reward machine}

\section{Introduction}
\subsection{Background of AI-Powered AutoPT}
In the rapidly advancing digital epoch, ensuring the security of information systems has emerged as a paramount concern. One of the effective methods used to evaluate the security of a computer system is Penetration Testing (PT)~\cite{engebretson2013basics}, which is important in probing for and identifying potential vulnerabilities that could be exploited by malicious entities. However, traditional PT requires highly specialized skills and domain expertise and can only be performed by trained professionals. Moreover, manual PT on a medium-sized LAN can be time-consuming and labor-intensive, sometimes lasting for several days or weeks~\cite{denis2016penetration}. Furthermore, PT can also cause substantial system downtime~\cite{ghanem2019reinforcement}. As a result, the security industry requires the development of automated PT techniques (AutoPT).

So far, many advanced AutoPT tools and frameworks have been created in the industry to increase PT efficiency. Metasploit~\cite{maynor2011metasploit}, for example, helps collect information and exploit vulnerabilities. However, the current automation level of these tools is relatively contextually limited (restricted to specific tasks) and unoptimized, preventing them from autonomously conducting comprehensive tests and evaluations on extensive assets, compared to the proficiency of human experts~\cite{10278377}.

One promising solution to advance PT performance is the application of reinforcement learning (RL) or deep RL (DRL). RL/DRL is a branch of artificial intelligence (AI) used to tackle sequential decision-making tasks. This involves an intelligent agent that takes actions within a specific environment, aiming to maximize its cumulative rewards over time through a process of trial and error~\cite{wong2023deep}.
RL/DRL has already shown impressive accomplishments in a range of artificial intelligence (AI) applications, such as autonomous driving, robotics, DeepMind AlphaGo, and OpenAI ChatGPT, among others~\cite{shao2019survey,kiran2021deep}. PT powered by RL/DRL can automate the process, improve PT efficiency, and have the potential to identify complex or hidden vulnerabilities (or critical attack paths) that a human tester might not consider~\cite{li2023innes}.

In recent years, there has been a rise in research on RL/DRL-powered PT on information systems. Schwartz et al.~\cite{schwartz2019autonomous} formulated the PT process as a Markov decision process (MDP) where the network configuration serves as the states and the available scans and exploits serve as the actions. The optimal PT policy with respect to the designed MDP is solved by using the deep Q-Network (DQN) algorithm. 
Similarly, Ryusei et al.~\cite{maeda2021automating} proposed a method to automate post-exploitation by combining DRL and the PowerShell Empire~\cite{powershellempire}. The results of the experiment show that the trained agent using the A2C algorithm could obtain administrative privileges from the domain controller.
To further minimize manual labor costs in PT, Ghanem et al. introduced the Intelligent Automated Penetration Testing System (IAPTS) developed based on the partially observable MDP (POMDP) formulation of PT~\cite{ghanem2018reinforcement,ghanem2019reinforcement}. This system combines an RL module with existing industrial PT frameworks. 
Hu et al.~\cite{hu2020automated} introduced a two-stage PT method based on DRL in communication networks. The first stage involves the use of scanning tools to collect network data and create an attack tree. In the second stage, the DQN algorithm is employed to determine the most efficient and impactful attack path from the attack tree. 
Tran et al.~\cite{tran2021deep} introduced a hierarchical DRL method for PT to address its large discrete action space, where a decomposition strategy for the action space is utilized. 
Qianyu et al.~\cite{li2023innes} proposed the INNES model for DRL-based PT to characterize its observation space and action space of PT, which makes the MDP formulation of PT more accurate. For PT in industrial control systems (ICSs), such as the power grid,~\cite{li2022deep} proposed to apply the DQN algorithm to identify the optimal ordering and timing of replay attacks on PMU packets that can trigger grid voltage violations.

\subsection{Challenges of AI-Powered AutoPT}
However, existing RL/DRL-based PT approaches still face the sampling efficiency issue in the training process, during which the agent needs a large number of agent-environment interactions to obtain the optimal policy.
One reason is because of the large action space, where the pen-tester has various attack actions to choose from.  According to MITRE ATT\&CK, a knowledge base of adversary tactics and techniques~\cite{attackmitre}, there are 14 attacking tactics (e.g., privilege escalation) for the Enterprise category. Each tactic has a set of techniques (e.g., access token manipulation), and each technique may have a set of sub-techniques. Additionally, each technique or subtechnique may also have a set of parameters to choose to form an atomic action for RL/DRL agents.

The difficulty of reward specification in RL/DRL-based PT is another challenge to consider. Previous studies assign 1) positive rewards to the agent when it successfully obtains network resources or takes impactful actions; and 2) negative rewards as penalties when the agent takes invalid actions or violates certain constraints. However, expressing these specification rules in a single reward function could increase its complexity, making it difficult for the agent to differentiate which aspects of contributions or losses result from its actions. Therefore, the agent may need many more interactions to figure it out. 

Moreover, interpretability is often lacking in RL/DRL-based PT. The trained PT policy cannot explicitly identify the current phase or situation of the PT agent and the subsequent direction it shall be heading. This type of perception or awareness could be encoded into the agent's neural networks representing the DRL policy through training, but it remains challenging to extract such information by decoding the neural networks.


\subsection{Contributions of This Work}
In response to the aforementioned challenges with respect to sampling efficiency, reward specification, and interpretability for RL/DRL-based PT, we propose to embed knowledge from the cybersecurity domain as guidelines into the agent's learning process in an explainable way. This approach can automatically break down the complex PT task into multiple subtasks, thereby enabling the agent to learn more efficiently. 

More specifically, we propose a knowledge-informed AutoPT framework called DRLRM-PT, which utilizes a reward machine (RM) to encode domain knowledge based on cybersecurity knowledge bases, such as MITRE ATT\&CK~\cite{attackmitre} and Cyber Kill Chain~\cite{cyberkillchain}, etc. The RM specifies a set of events that occurred in PT and decomposes PT into multiple subtasks based on existing PT practices. Moreover, RM can specify different reward functions for PT in different phases, expanding the flexibility of traditional reward functions in RL/DRL-based PT. Under the DRLRM-PT framework, PT is formulated as a POMDP guided by RMs. We focus specifically on lateral movement as a case study, which assumes the pen-tester has gained the initial access to the network and will move deeper into the network for owning high-value assets. We investigate two different RMs as two guidelines. Finally, we employ the deep Q-learning algorithm with RM (DQRM) to solve the POMDP and optimize the PT policy. 

To the best of our knowledge, this work is the first attempt to integrate the domain-specific knowledge and expertise of humans into DRL to automate and enhance PT in complex network systems. In brief, the contributions of this work can be summarized as follows:
\begin{itemize}

\item We proposed a knowledge-informed AutoPT framework (DRLRM-PT), which utilizes RMs to encode domain knowledge as guidelines to train PT policies.

\item We applied RMs to specify and arrange subtasks of PT based on existing PT practices. We also designed a dedicated event set for AutoPT , where events represent successful executions of adversary tactics defined in cybersecurity knowledge bases, which are used by RMs to assign subtasks to agents during PT.

\item We took the lateral movement of PT as a case study and formulated PT as a POMDP guided by RMs, where two RMs were designed based on MITRE ATT\&CK knowledge base; DQRM algorithm was applied to optimize the PT policy. This case study proves the effectiveness of our proposed approach and provides an illustrative example of embedding existing cybersecurity knowledge bases into RL/DRL-based PT.
\end{itemize}

This paper is organized as follows: Section~\ref{section-DRLRM-PT} presents the proposed DRLRM-PT framework. Section~\ref{POMDPRM-design} introduces the POMDP formulation with RM specifically for the lateral movement of PT. Section~\ref{simulation-platform} describes the simulation platform and testing environments. Comparative studies are performed to validate the PT performance of DQRM agents In Section~\ref{experimental-validation}. The conclusions and future works are discussed in Section~\ref{conclusion}.

\section{Knowledge-Informed AutoPT Framework}\label{section-DRLRM-PT}

\subsection{DRLRM-PT Framework}

In this study, we investigate an AutoPT method, which employs a computer program acting as an agent to launch a series of cyberattacks on network systems. The objective is to discover possible attack paths to take ownership of critical resources, similar to the capture-the-flag game. Thus, PT is a sequential decision-making problem, which can be formulated as a partially observable Markov decision process (POMDP)~\cite{ghanem2019reinforcement}, and the optimal PT policy with respect to the POMDP can be obtained by applying RL/DRL algorithms. 

\begin{figure}[b] 
\centering
\includegraphics[width=1.0\columnwidth]{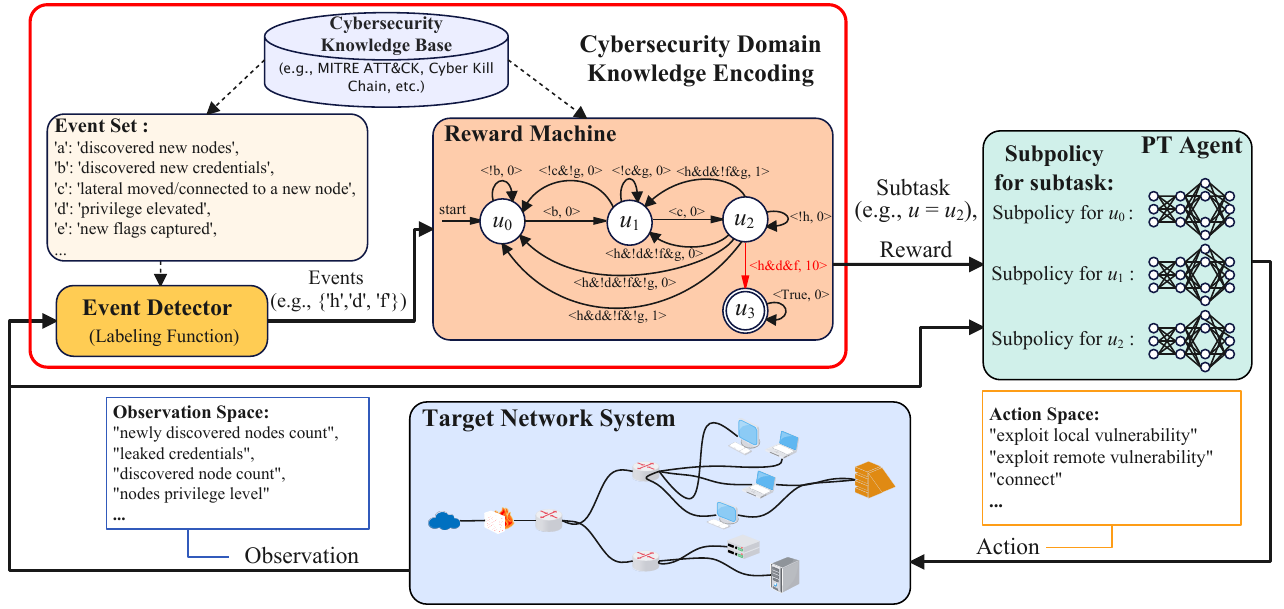}
\caption{The proposed knowledge-informed AutoPT framework (DRLRM-PT).}
\label{DRLRM-PT-framework}
\end{figure}

To address the challenges of sampling efficiency, reward specification, and interpretability in RL/DRL-based PT, we propose a knowledge-informed AutoPT framework based on RL with reward machine (DRLRM-PT), as depicted in Figure~\ref{DRLRM-PT-framework}, which uses a reward machine (RM) to encode domain knowledge of PT based on existing cybersecurity knowledge bases.

This framework involves an agent that acts as a pen-tester and interacts with the target network system that makes up the environment. This environment typically consists of hosts, firewalls, routers, and communication channels, among other components. The agent's action is selected from a range of PT activities (action space), such as network scanning, exploiting vulnerabilities, lateral movement, and privilege escalation. Additionally, the agent can gather observations from the environment through the information from scanning operations, defined by an observation space containing newly discovered vulnerabilities, leaked credentials, etc.
The immediate reward reflects the evaluation of the agent's action. The aim of PT typically involves taking ownership of critical resources within the networks, such as customer data. Thus, the agent can assign a positive reward to itself when the action is beneficial for accomplishing this goal or a negative reward when it is not. The agent wants to find an optimal PT policy to maximize accumulated rewards through learning experiences from agent-environment interactions~\cite{sutton2018reinforcement}. To tackle the challenge of high dimensionality of the target system arising from the large observation or action spaces\cite{nguyen2019deep}, the PT policy is represented by deep neural networks to determine the best action based on input observation. During the training phase, after each step, the agent will use the experience (observation, action, reward, next observation) to update the weights of the neural networks (iteratively improving the PT policy).

In this framework, the agent is informed and guided by a cybersecurity domain knowledge encoding module, which uses an RM to encode the domain knowledge of PT based on the cybersecurity knowledge bases, such as MITRE ATT\&CK, Cyber Kill Chain, etc. 

The RM is a state machine with two essential functions: 1) decomposing the PT task into a series of subtasks, such as ``discover credentials" and ``privilege escalation", and organizing them using a state machine structure; 2) specifying a reward function for each state transition within the RM. 

The RM receives a set of events detected during PT as input, leading to a transition of its internal state from one state to another according to its transition rules (logic formula over the event set). The output of the RM includes the RM state and a reward function. The RM state indicates the completion of the previous subtask and the initiation of a new subtask; thus, it can be used to denote the subtask ID.

The event detector (also called the labeling function under the RM theory~\cite{icarte2018using}) can identify occurring events, denoted by symbols, e.g., `a' for ``discovered new nodes" and `b' for ``discovered new credentials", from the environment by analyzing the latest agent-environment interaction. These events are designed to represent successful executions of adversary tactics as defined by cybersecurity knowledge bases. Because these tactics serve as the adversary's tactical goals, providing the motivation behind their actions~\cite{attackmitre}. For instance, privilege escalation is one tactic for attacking enterprise networks defined by ATT\&CK. Therefore, the corresponding event can be defined as its successful outcome, i.e., ``privilege escalated". All events are predefined in an event set and can be detected using the event detector. Table~\ref{AP} shows an example of an event set designed for PT based on ATT\&CK tactics. 
\begin{table}[h]
\caption{Event Set of PT ($\mathcal{P}$)}
\vspace{-3mm}
\begin{center}
\begin{tabular}{c c }
\hline
\specialrule{0em}{1pt}{1pt}
\textbf{Event Symbol} &\textbf{Event Description}\\
\specialrule{0em}{1pt}{1pt}
\hline
\specialrule{0em}{1pt}{1pt}

`a' & Discovered new nodes\\ 
\specialrule{0em}{1pt}{1pt}

`b' & Discovered new credentials\\ 
\specialrule{0em}{1pt}{1pt}

`c' & Lateral moved (connected to a new node)\\ 
\specialrule{0em}{1pt}{1pt}

`d' & Privilege elevated\\ 
\specialrule{0em}{1pt}{1pt}

`e' & New flags captured (new target nodes owned)\\ 
\specialrule{0em}{1pt}{1pt}

`f' & Achieved the PT goal \\ 
\specialrule{0em}{1pt}{1pt}

`g' & Has unused credentials \\ 
\specialrule{0em}{1pt}{1pt}

`h' & Taken action to elevate the privilege \\ 
\specialrule{0em}{1pt}{1pt}
\specialrule{0em}{1pt}{1pt}

\hline
\end{tabular}
\label{AP}
\end{center}
\end{table}

As an example shown in Figure~\ref{DRLRM-PT-framework}, when events `h', `d', and `f' are detected, the RM will transit its state from $u_2$ to $u_3$ and output $u_3$ (subtask ID) and a constant reward function (reward = 10) according to its defined transition rule $<\text{h} \& \text{d} \& \text{f}, 10>$. 

In addition, RM outputs a reward function instead of a reward value. Thus, RM allows the agent to specify task-specific reward functions to enhance the flexibility of the single reward function used in traditional RL algorithms. 

Furthermore, DRLRM-PT utilizes deep Q-learning with RM algorithm (DQRM) to train a PT policy, which decomposes the policy into a set of sub-policies for every subtask and can train all sub-policies simultaneously. Therefore, the final PT policy actually selects the sub-policy to determine the PT action. 


\subsection{POMDP with Reward Machine Formulation for PT}\label{subsection-POMDPRM}

The proposed PT under the DRLRM-PT framework is formulated as a POMDP with RM, which is characterized by a tuple consisting of seven components $\left \langle \mathcal{S}, \mathcal{A}, \mathcal{O}, T(s_{t+1} | s_t, a_t), O(o_{t} | s_t), \mathcal{R},\gamma \right \rangle$~\cite{toro2021learning}. 

$\mathcal{S}$ represents the set of environment states, where $s_t \in \mathcal{S}$ denotes the state at time $t$. $\mathcal{O}$ represents the set of observations, and $o_t \in \mathcal{O}$ denotes the observation at time $t$. $\mathcal{A}$ represents the set of actions, and $a_t \in \mathcal{A}$ denotes the action taken by the agent at time $t$. $T(s_{t+1} | s_t, a_t)$ is the probability of transition from the current state $s_t$ to the next state $s_{t+1}$ when the agent performs an action $a_t$. $O(o_{t} | s_t)$ is the emission function that maps the current state to the observation received by the agent. 

$\mathcal{R}$ is the RM, which is a tuple with six components $\mathcal{R} = \left \langle \mathcal{P}, L, U, u_0, \delta^u, \delta^r \right \rangle$. 
$\mathcal{P}$ is an event set of PT. 
$L$ is the labeling function (event detector), $L: \mathcal{O} \times \mathcal{A} \times \mathcal{O} \rightarrow 2^{\mathcal{P}}$, which assigns truth values to events (True: event occurred, False: event not occurred), by analyzing the input experience ($o_t, a_t, o_{t+1}$). 
$U$ is a finite set of RM states. $u_0 \in U$ is an initial state. 
$\delta^u$ is the state-transition function, $\delta^u: U \times 2^{|P|} \rightarrow U$, which determines the next RM state based on its current state and the captured events (i.e., $u_{t+1} \leftarrow \delta^u(u_t, L(o_{t}, a_t, o_{t+1}))$).
$\delta^r$ is the reward-transition function,
$\delta^r: U \times 2^{|P|} \rightarrow [\mathcal{O} \times \mathcal{A} \times \mathcal{O} \rightarrow \mathbb{R}]$, which outputs a reward function based on its current state and the captured events (i.e., $R(o_t, a_t, o_{t+1}) \leftarrow \delta^r(u_t, L(o_{t}, a_t, o_{t+1}))$). The agent can use the output reward function to obtain the reward. 

$\gamma \in [0,1)$ is the discount factor that determines the trade-off between immediate and long-term rewards that the agent prefers to achieve.


Due to the complex nature of the target network system, the determination of $T(s_{t+1} | s_t, a_t)$ and $O(o_{t} | s_t)$ poses challenges for PT. However, the agent can take the environment as a black box and learn the policy through pure trial and error. 

For a specific scope of PT, the action space, observation space, and RM can be customized accordingly. 

\section{POMDP with RM Design for Lateral Movement}\label{POMDPRM-design}

In this work, the lateral movement on enterprise networks is considered as the study case of PT, which is under the assumption that the agent has already entered the target network (post-exploitation assumption). The POMDP with RM formulation and two RMs are designed in the following subsections.

\subsection{Action Space}
We consider three types of actions that the agent can execute during lateral movement. The first is scanning, which involves collecting network information by discovering new machines (nodes), determining the connections between these nodes, acquiring machine configuration data, and gathering vulnerability information for discovered nodes. 

The second type of action is vulnerability exploitation, which can be classified into local vulnerability exploitation and remote vulnerability exploitation. Local vulnerability exploitation can only be performed on a connected node (the node where the agent is operating), and the agent seeks to steal local information, increase host privileges, or discover credentials for connecting to other nodes. Remote vulnerabilities come from nodes that are currently discovered but are not owned by the agent. By exploiting remote vulnerabilities, the agent can gather more information about the remote nodes. 

The third type of action is connection, which enables the agent to connect a node using specific credentials and ports.

Due to the agent's lack of direct access to action outcomes, scanning is considered a mandatory action that must be performed after each action rather than being optional in action space. Additionally, the scanning operation can also contribute to forming an observation.

\begin{table}
\caption{Action space of PT}
\begin{center}
\begin{tabular}{c c c}
\hline
\specialrule{0em}{1pt}{1pt}
\textbf{Action} &\textbf{Notation} & \textbf{Subspace Size}\\
\specialrule{0em}{1pt}{1pt}
\hline
\specialrule{0em}{1pt}{1pt}

Local vulnerability exploit &$[i, l]$ & $\hat{n} \times \hat{n}_l$ \\ 
\specialrule{0em}{1pt}{1pt}

Remote vulnerability exploit &$[i, j, r]$ & $\hat{n} \times (\hat{n}-1) \times \hat{n}_r$\\ 
\specialrule{0em}{1pt}{1pt}

Connection &$[i, j, p, c]$ & $\hat{n} \times (\hat{n}-1) \times \hat{n}_p \times \hat{n}_c$ \\ 
\specialrule{0em}{1pt}{1pt}

\hline
\end{tabular}
\vspace{-6mm}
\label{action-space}
\end{center}
\end{table}

The action space is listed in Table~\ref{action-space}, which includes three subspaces for local vulnerability exploitation, remote vulnerability exploitation, and connection, respectively. $i$, $j$, $l$, $r$, $p$, and $c$ denote the ID of the source node, the target node, the local vulnerability, the remote vulnerability, the port, and the credentials, respectively. $\hat{n}$, $\hat{n}_l$, $\hat{n}_r$, $\hat{n}_p$, and $\hat{n}_c$ are agent's estimations of the maximum number of nodes, local vulnerabilities, remote vulnerabilities, ports, and credentials, respectively. Table~\ref{action-space} implies that for local vulnerability exploitation, the agent should choose the node ID and local vulnerability ID. For remote vulnerability exploitation, the agent should choose the source node ID, target node ID, and remote vulnerability ID. The source node ID, target node ID, port ID, and credential ID should be set for the connection. The PT action is a vector selected from one of the subspaces. 

We consider each action will last a constant unit period, after which it will be considered completed and terminated.

\subsection{Observation Space}
The observation of the agent in PT is obtained by scanning operation after each action taken using scanning tools, such as Nmap~\cite{orebaugh2011nmap}. The observation space is designed and shown in Table~\ref{observation-space}, which consists of many subspaces, including the \textit{discovered nodes count}, \textit{nodes privilege level, discovered nodes properties}, \textit{leaked credentials}, and \textit{lateral move}. 

\begin{table}[h]
\caption{Observation space of PT}
\vspace{-3mm}
\begin{center}
\begin{tabular}{c c c}
\hline
\specialrule{0em}{1pt}{1pt}
\textbf{Observation} &\textbf{Notation} & \textbf{Subspace Size}\\
\specialrule{0em}{1pt}{1pt}
\hline
\specialrule{0em}{1pt}{1pt}

Discovered nodes count &$n_d$ & $\hat{n}$ \\ 
\specialrule{0em}{1pt}{1pt}


Nodes privilege level &$[a_i]_{\hat{n}}$     &$2^{\hat{n}}$ \\ 
\specialrule{0em}{1pt}{1pt}

Discovered nodes properties &$[a_{i,p}]_{\hat{n} \times \hat{n}_{pr}}$  &$3^{\hat{n} \times \hat{n}_{pr}}$ \\ 
\specialrule{0em}{1pt}{1pt}

Leaked credentials &$[a_{i,p,c}]_{\hat{n} \times \hat{n}_{p} \times \hat{n}_c}$  &$3^{\hat{n} \times \hat{n}_{p} \times \hat{n}_c}$\\ 
\specialrule{0em}{1pt}{1pt}

Lateral move &$b_l$  & $2$\\ 
\specialrule{0em}{1pt}{1pt}

\hline
\end{tabular}
\label{observation-space}
\end{center}
\end{table}

Among them, \textit{nodes privilege level} is a vector describing the privilege level of every node, where two values can be assigned to each entry: 0 = ``not owned", 1 = ``Admin". 

\textit{Discovered nodes properties} tells what properties each node has. Properties include different types of operating systems (e.g., Windows, Linux),  different types of databases (e.g., SQLServer, MySQL), etc. It is represented by a $\hat{n} \times \hat{n}_{pr}$ matrix, where $\hat{n}_{pr}$ is the estimated maximum number of properties. Each entry has three values: 0 = ``No", 1 = ``Yes", 2 = ``Unknown". 

\textit{Leaked credentials} is a $\hat{n} \times \hat{n}_{p} \times \hat{n}_c$ tensor with the first dimension indicating the target node ID, the second dimension indicating the port ID, the third dimension indicating the credential ID. Each entry has three values: 0 = ``Not discovered", 1 = ``Used", 2 = ``Unused".

\textit{Lateral move} indicates whether the agent successfully moves from one node to another in a new interaction with two values: 0 = ``No", 1 = ``Yes". 

By flattening and concatenating all vectors from these five subspaces, we can get an observation vector.



\subsection{Reward Machine I ($\mathcal{R}_1$)}

In this work, the PT agent is guided by the domain knowledge of PT encoded in an RM.  
In the field of PT, one useful guideline is that the pen-tester attempts to discover as many login credentials as possible to gain access and control over as many nodes as possible. Therefore, the PT can be divided into three subtasks: 1) discover new credentials, 2) gain access (connect) to a new node by using the discovered credentials, and 3) elevate the privilege of the connected node to own its properties. This process will be repeated to own more and more nodes until the PT goal is met, such as discovering critical data or owning a specific number of nodes.

\begin{figure}[htbp] 
\centering
\includegraphics[width=0.65\columnwidth]{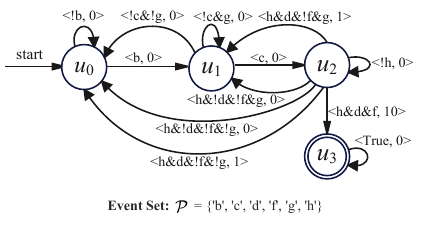}
\caption{The diagram of Reward Machine I ($\mathcal{R}_1$).}
\label{RM-1}
\end{figure}

According to this guideline, our first designed RM ($\mathcal{R}_1$) is shown in Figure~\ref{RM-1}, which has four states ($U_1 = \{u_0, u_1, u_2, u_3\}$) starting from $u_0$ and terminates at $u_3$. Its event set $\mathcal{P}_1$ = \{`b', `c', `d', `f', `g', `h'\} is a subset of the set defined in Table~\ref{AP}.

$\mathcal{R}_1$ tells that the agent will stay on $u_0$ until new credentials are found (`b'), then transition to $u_1$. The agent will remain in $u_1$ if no successful connection is made (`c') and the previously discovered credentials do not run out (`g'). If all credentials are used (`!g'), but still cannot connect to a new node (`!c'), it will go back to $u_0$ to find new credentials. If the agent can connect to a new node from $u_1$ (`c'), it will move to $u_2$. Then, the agent tries to take actions to elevate the privilege level of the connected node (`h'). If the privilege level is elevated (`d') and the final goal is reached (`f'), $\mathcal{R}_1$ ends in $u_3$. If the final goal is not reached (`!f'), the agent checks for unused credentials. If it is (`g'), the agent returns to $u_1$ to try other credentials. If not, the agent returns to $u0$ to find new credentials.

Each transition in $\mathcal{R}_1$ also outputs a reward value. According to Figure~\ref{RM-1}, when the privilege level of a connected node is elevated (`d'), the agent gets a reward = 1. When the final goal is reached (`f'), the agent gets a reward = 10.

\subsection{Reward Machine II ($\mathcal{R}_2$)}

\begin{figure}[htbp] 
\centering
\includegraphics[width=0.65\columnwidth]{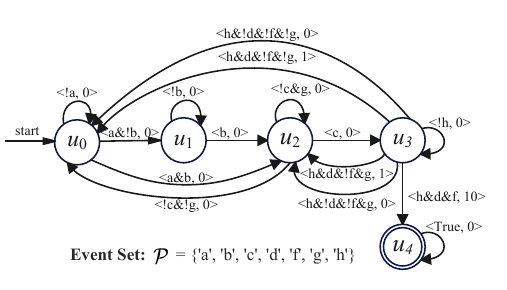}
\caption{The diagram of Reward Machine II ($\mathcal{R}_2$).}
\label{RM-2}
\end{figure}

We also investigate a more detailed RM $\mathcal{R}_2$ where the knowledge guides the agent to discover new nodes first, then discover new credentials. Next, connect to a new node, and finally, elevate the privilege of the connected node. Therefore, its event set is $\mathcal{P}_2$ = \{`a', `b', `c', `d', `f', `g', `h'\}, where `a' is added. The diagram of $\mathcal{R}_2$ is shown in Figure~\ref{RM-2}. $\mathcal{R}_2$ has five states since it has one more task compared to $\mathcal{R}_1$. 

According to Figure~\ref{RM-2}, the agent will stay on $u_0$ until new nodes are discovered (`a'), then transition to $u_1$. Sometimes, the agent also finds credentials simultaneously (`a\&b') because discovering a new credential usually comes with a new target node; thus, it can directly transition to $u_2$. The transitions in $\mathcal{R}_2$ from $u_1$ are similar to those in $\mathcal{R}_2$ from $u_0$. One difference is that after a successful connection (`c') if all discovered credentials are run out (`!g') but still cannot achieve the PT goal (`!f'), it will transition to $u_0$ to find new nodes instead of to find new credentials. 

The reward transition function of $\mathcal{R}_2$ is similar to that of $\mathcal{R}_1$. When the connected node's privilege level is elevated (`d'), the agent receives a reward of 1. When the agent reaches the final goal (`f'), the reward is 10.

\subsection{PT Objective}
The goal of lateral movement is to take ownership (elevate privilege) of as many nodes as possible to own their properties. Therefore, discounted accumulated rewards with respect to $\mathcal{R}$ (denoted by $G_{\mathcal{R}}$) during PT can be used as the objective function for the agent to maximize, as expressed in Eq.~\ref{objective function}.

\begin{equation}
\label{objective function}
G_{\mathcal{R}} = \sum_{t=1}^{N_{it}} \gamma^{t-1} r_t,
\end{equation}
where $N_{it}$ is the total number of actions taken in the PT. $r_t$ is the immediate reward obtained by $\mathcal{R}$. $\gamma$ is the discount factor. By optimizing Eq.~\ref{objective function}, we hope to find out an optimal PT policy with respect to $\mathcal{R}$, denoted as $\pi^{*}_{\mathcal{R}} = \arg\max_{\pi} G_{\mathcal{R}}$.

\subsection{PT policy optimization based on DQRM}

The deep Q-learning with RM algorithm (DQRM) is used to train the agent and obtain the optimal PT policy $\pi^{*}_{\mathcal{R}}$. It is an updated version based on the Q-learning with RM (QRM).  

Give an RM $\mathcal{R}$, QRM decomposes the training process by learning one Q-function ($Q(o,a)$, a function used to evaluate future rewards for a certain observation and action) per state in $\mathcal{R}$~\cite{icarte2022reward}. These Q-functions can be considered as subpolicies and represented by Q-tables. We denote $Q_u$ as the Q-function for state $u$ of $\mathcal{R}$. Since the terminal state will bring no future rewards, the corresponding Q-function will always output 0. For example, in Figure~\ref{RM-1}, $\mathcal{R}_1$ has four states and will have four Q-functions. 

As the complexity of the environment increases, the observation space of PT defined in Table~\ref{action-space} will grow exponentially, bringing a curse of dimensionality challenging to the QRM. Therefore, we adopt DQRM, which utilizes deep neural networks (called Q-networks and are parameterized by $\theta_u$) instead of Q-tables to approximate Q-functions. It also adopts target Q-networks (parameterized by $\theta^-_u$) and the experience replay mechanism to stabilize the training process. 

The update rule for Q-networks is as follows:
\begin{equation}
\label{update-rule-DQRM}
Q_u(o,a; \theta_u) \stackrel{\alpha}{\longleftarrow}
r(o,a,o') + \gamma \max_{a'} Q^-_{u'}(o',a'; \theta^-_{u'}),
\end{equation}
where $Q_u$ is the Q-network of state $u$; $Q^-_{u'}$ is the target Q-network of the next state $u'$. 

These Q-networks will be iteratively updated through the agent's interactions with the environment until the end of training. Finally, the optimal action $a^*_t$ given $o_t$ can be obtained by:
\begin{equation}
\label{a_optimal}
a^*_t = \arg\max_{a \in\mathcal{A}}  Q_u(o_t, a; \theta_u).
\end{equation}

The pseudo-code of the PT optimization using DQRM is shown in Algorithm~\ref{algorithm}.

\begin{algorithm}
\small
\caption{PT policy optimization using DQRM}
\label{algorithm}
\begin{algorithmic}[1]
\Require $\mathcal{A}, \mathcal{O},\gamma, \alpha, \mathcal{R} = \left \langle \mathcal{P}, L, U, u_0, \delta^u, \delta^r \right \rangle, C$
\State For every $u \in U-\{u_{T}\}$, initialize $Q_u(\cdot)$ and $Q^-_u(\cdot)$ with parameters $\theta_u$ and $\theta^-_{u}$, respectively;
\State Initialize an experience replay buffer $D$;
\For{$episode$ = 1 to $N_{ep}$} 
    \State Agent takes scanning operation to get the initial observation $o_0$ from the target network system, $o_t \leftarrow o_0$;
    \State $u_t \leftarrow u_0$;
    \For{$t$ = 1 to $N_{it}$} 
    \State Agent selects an action $a_t$ under $o_t$ from $\mathcal{A}$ based on  $\epsilon$-greedy strategy; 
    \State Agent executes $a_t$ and takes scanning operation to get the observation $o_{t+1}$ from the target network system;
    \State $u_{t+1} \leftarrow \delta^u(u_t, L(o_{t}, a_t, o_{t+1}))$;
    \State $R(\cdot) \leftarrow \delta^r(u_t, L(o_{t}, a_t, o_{t+1}))$;
    \State $r_{t+1} = R(o_{t}, a_t, o_{t+1})$;
     \State Save the experience $(o_t, a_t, o_{t+1}, u_t, u_{t+1}, r_{t+1})$ into $D$;
    \State Sample random mini-batch of experiences $(o, a, o', u, u', r)$ from $D$, and update $\theta_u$ for $Q_u(\cdot)$ based on Eq.~\ref{update-rule-DQRM};
    \If{$t$ mode $C$ = 0 ?}
    \State For every $u \in U-\{u_{T}\}$, $\theta^-_u \leftarrow \theta_u$;
    \EndIf
    \State $o_t \leftarrow o_{t+1}$
    \State $u_t \leftarrow u_{t+1}$
    \EndFor
\EndFor
\end{algorithmic} 
\end{algorithm}

\section{Simulation Platform and Testing Environments}\label{simulation-platform}
In this study, we use \textit{CyberBattleSim} as our experimental platform~\cite{cyberbattlesim}. This platform, developed by Microsoft, is an open-source network simulator designed for lateral movement research. It enables users to examine PT strategies they establish on simulated enterprise networks. Simulation networks are modeled by an abstract graph with nodes (machines) parameterized by a set of pre-planted vulnerabilities. 
In addition, the OpenAI Gym interface allows for the training of automated agents by utilizing RL/DRL algorithms.

\begin{figure}[htbp] 
\centering
\includegraphics[width=1\columnwidth]{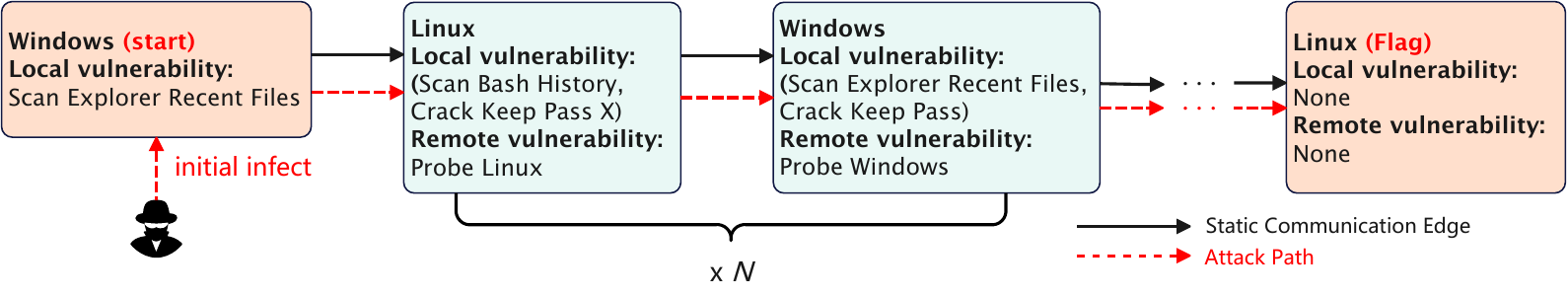}
\caption{\textit{CyberBattleChain} environment (\textit{env}-1).}
\label{env-chain}
\end{figure}

\begin{figure}[htbp] 
\centering
\includegraphics[width=1\columnwidth]{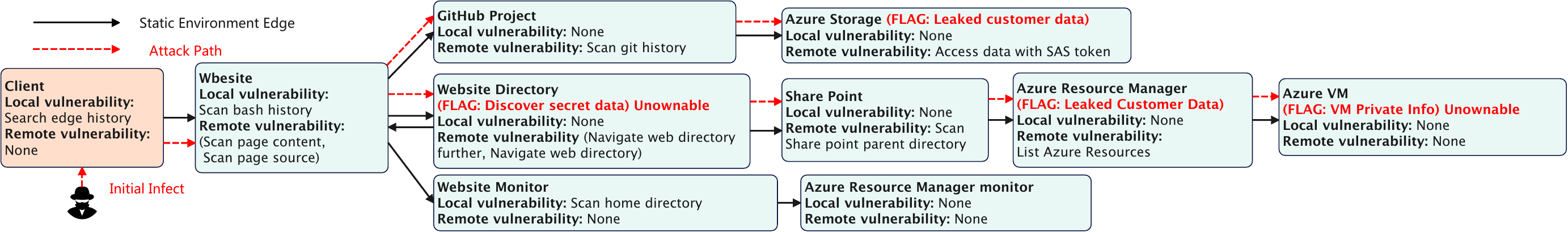}
\caption{\textit{CyberBattleToyCtf} environment (\textit{env}-2).}
\label{env-ctf}
\end{figure}

In \textit{CyberBattleSim}, two typical networks are set as selectable environments to test PT strategies, called \textit{CyberBattleChain} (denoted as \textit{env}-1) and \textit{CyberBattleToyCtf} (denoted as \textit{env}-2), as shown in Figure~\ref{env-chain} and \ref{env-ctf}, respectively. Each node in both environments possesses a set of properties and is pre-planted with a set of local and remote vulnerabilities. These vulnerabilities may reveal the credentials of adjacent nodes or lead to the privilege escalation of target nodes. Moreover, both environments are pre-configured with `flag' nodes, which contain important or sensitive resources, such as customer data. The agent's goal is to capture as many flags as possible during the PT while minimizing the number of actions.

\textit{env}-1 has a sequential network structure with the flag located at the end node, as shown in Figure~\ref{env-chain}. \textit{N} means the number of ``Linux-Windows" link. In this case, we set \textit{N} to 8. Therefore, to capture the flag, the agent needs to elevate the privileges of all nodes sequentially.

Compared to \textit{env}-1, \textit{env}-2 is more complex.  \textit{env}-2 is made up of a mesh network architecture, as shown in Figure~\ref{env-ctf}.  \textit{env}-2 contains four flags, of which two are unattainable since they did not leak credentials and the attacker cannot gain higher privileges, but the attacker can discover other nodes from it by performing remote attacks, such as navigating the web directory. The agent's goal is to capture two attainable flags.

\section{Experimental Validation}\label{experimental-validation}
We design the experiments to validate our proposed method and attempt to answer two research questions below:
\begin{itemize}
\item \textbf{RQ1}: Can the agent guided by RM improve the learning efficiency of PT compared to the agent without RM?
\item \textbf{RQ2}: How will different RM designs affect the PT performance. 
\end{itemize}

\subsection{Agent Configurations}


Table~\ref{Training_agents} lists four different agents, where DQRM-RM1 and DQRM-RM2 are agents that use the DQRM algorithm with $\mathcal{R}_1$ and $\mathcal{R}_2$, respectively. DQN-RM1 and DQN-RM2 are agents that use the DQN algorithm where only the reward is obtained by $\mathcal{R}_1$ and $\mathcal{R}_2$, respectively. 

\begin{table}[ht]
\caption{Learning Agents}
\begin{center}
\begin{tabular}{c c}
\hline
\specialrule{0em}{1pt}{1pt}
\textbf{Agent Name} &\textbf{Description}\\
\specialrule{0em}{1pt}{1pt}
\hline
\specialrule{0em}{1pt}{1pt}

DQRM-RM1 & DQRM guided and rewarded by $\mathcal{R}_1$   \\ 
DQN-RM1  & DQN rewarded by $\mathcal{R}_1$   \\
DQRM-RM2 & DQRM guided and rewarded by $\mathcal{R}_2$   \\
DQN-RM2  & DQN rewarded by $\mathcal{R}_2$   \\
\specialrule{0em}{1pt}{1pt}

\hline
\end{tabular}
\label{Training_agents}
\end{center}
\end{table}


The agents' training parameters are empirically determined, as listed in Table~\ref{Training-parameters}. 
We run 100 episodes to train each agent and 50 episodes to evaluate their trained policies. Each episode is an execution of one PT, which is terminated upon the achievement of the predefined PT goal (capture all flags) or the maximum number of actions performed.
The Q-functions are fully connected neural networks with two hidden layers, and each hidden layer has 150 neurons.
After each step, the agents will randomly sample 100 experiences (batch size) to update their policies. For every 10 steps, the target neural networks will copy the weights from the training neural networks. The learning rate of neural networks is set to 0.001. $\gamma$ is set to 0.9. $\epsilon$ is set to 0.3. 

\begin{table}[ht]
\caption{Training Parameters For DQRM and DQN}
\begin{center}
\begin{tabular}{c c c}
\hline
\specialrule{0em}{1pt}{1pt}
\textbf{Parameter} &\textbf{Description} & \textbf{Value}\\
\specialrule{0em}{1pt}{1pt}
\hline
\specialrule{0em}{1pt}{1pt}

$N_{it}$ & Maximum number of actions per episode & 1500   \\ 
\specialrule{0em}{1pt}{1pt}

$N_{ep}$ &  Number of episodes (train)&100 \\ 
\specialrule{0em}{1pt}{1pt}

$N_{evl}$ & Number of episodes (evaluation)&50  \\ 
\specialrule{0em}{1pt}{1pt}

$\gamma$ & Discount factor &0.9  \\ 
\specialrule{0em}{1pt}{1pt}

$\alpha$ &  \tabincell{c}{Learning rate}  &0.001  \\ 
\specialrule{0em}{1pt}{1pt}

$C$  & Target networks update frequency &10  \\ 
\specialrule{0em}{1pt}{1pt}

$N_{bch}$ & Batch size &100  \\ 
\specialrule{0em}{1pt}{1pt}

$N_{hl}$ &Number of hidden layers &2  \\ 
\specialrule{0em}{1pt}{1pt}

$N_{hs}$ &Hidden size &150  \\ 
\specialrule{0em}{1pt}{1pt}

$\epsilon$ & $\epsilon$-greedy parameter & 0.3 \\
\specialrule{0em}{1pt}{1pt}
\hline
\end{tabular}
\label{Training-parameters}
\end{center}
\end{table}

These four agents in Table~\ref{Training_agents} are comparable in the same environment since agents always receive the same reward given the same agent-environment interaction. 
The performance indicators of agent include training efficiency and PT efficiency, which will be compared in the training phase and in the evaluation phase, respectively. 
Training efficiency indicates how quickly an agent learns to achieve a high accumulated reward within a certain number of training episodes, which will be visualized by the accumulated reward with respect to the number of steps (actions) taken. PT efficiency indicates how quickly an agent achieves its goal during PT, which will be evaluated by the distribution of the total number of steps per episode among a certain number of evolution episodes.

\subsection{Comparative Studies}
\begin{figure}[bt] 
\centering
\includegraphics[width=0.65\columnwidth]{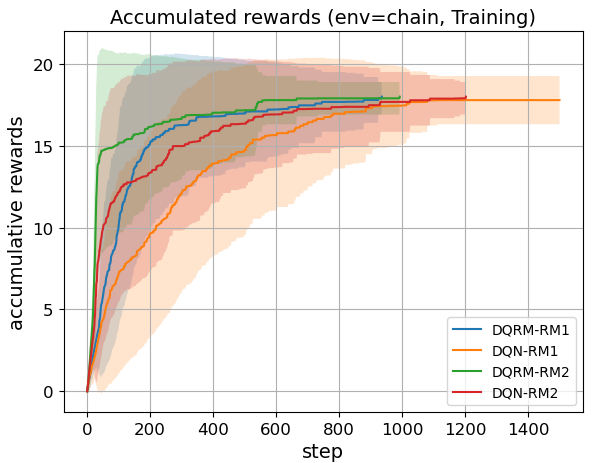}
\vspace{-2mm}
\caption{The training performance of four agents in \textit{env}-1.}
\vspace{-6mm}
\label{accumulated-rewards-training-chain}
\end{figure}

We train these four agents based on the configurations in \textit{env}-1 and \textit{env}-2 and plot results in Figure~\ref{accumulated-rewards-training-chain}, \ref{accumulated-rewards-evaluation-chain}, \ref{attack-steps-chain}, and \ref{attack-steps-ctf}. 

Figure~\ref{accumulated-rewards-training-chain} and \ref{accumulated-rewards-evaluation-chain} show the accumulated rewards with respect to the number of steps in \textit{env}-1 under the training phase and the evaluation phase, respectively. Figure~\ref{attack-steps-chain} shows the distributions of the total number of steps taken per episode in \textit{env}-1 for different agents, where the left side shows the training phase and the right side shows the evaluation phase. Similarly, Figure~\ref{attack-steps-ctf} presents the distributions of the total number of steps taken per episode in \textit{env}-2 for different agents in the training phase and evaluation phase.

\subsubsection{Comparison between DQRM agents and DQN agents} To answer \textbf{RQ1}, we compare the PT performance between the DQRM agents and the DQN agents in \textit{env}-1 and \textit{env}-2, guided by $\mathcal{R}_1$ and  $\mathcal{R}_2$, respectively. 

From Figure~\ref{accumulated-rewards-training-chain}, we can see that both DQRM-RM1 and DQRM-RM2 outperform DQN-RM1 and DQN-RM2 in terms of the training efficiency in \textit{env}-1, respectively. This is because by using the same number of actions, DQRM agents can obtain higher average accumulated rewards than DQN agents. As a result, the curve that exhibits superior performance will tend to approach the top left corner more closely. For instance, by 200 steps, the average accumulated rewards of DQRM-RM1 and DQRM-RM2 are about 15 and 16, respectively, whereas the average accumulated rewards of DQN-RM1 and DQN-RM2 are about 10 and 13, respectively.

\begin{figure}[ht] 
\centering
\includegraphics[width=0.7\columnwidth]{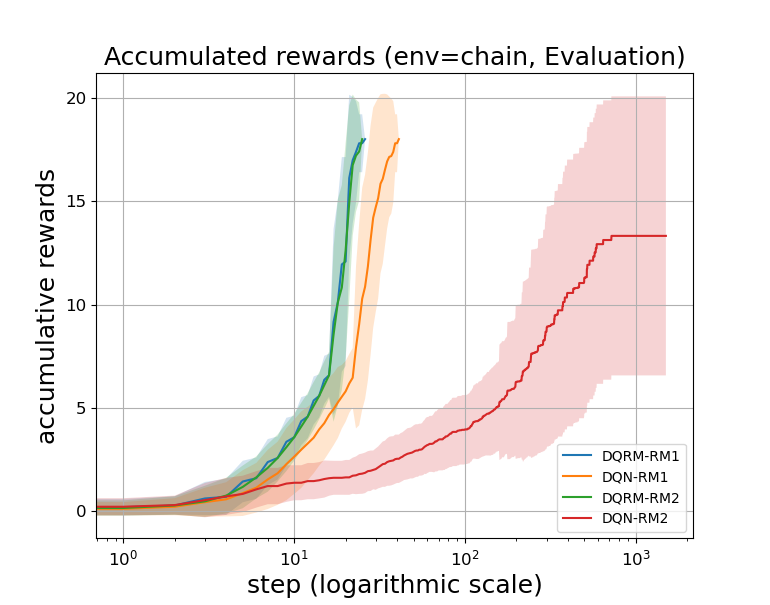}
\vspace{-2mm}
\caption{The evaluation performance of four agents in \textit{env}-1.}
\vspace{-4mm}
\label{accumulated-rewards-evaluation-chain}
\end{figure}

\begin{figure}
\centering
\includegraphics[width=0.65\columnwidth]{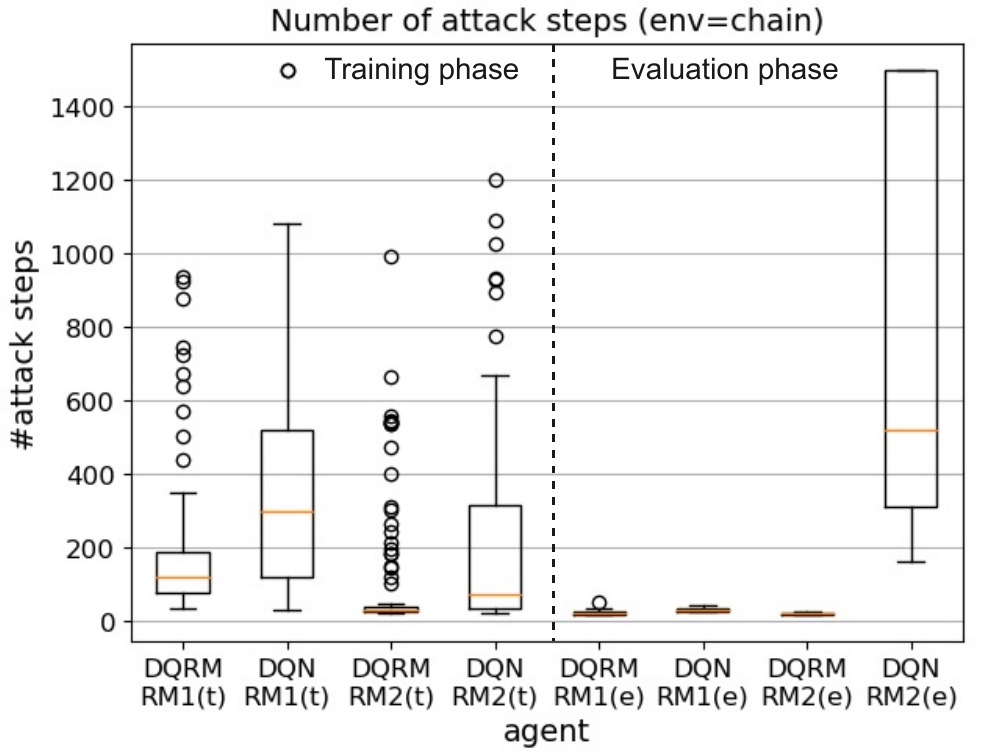}
\vspace{-3mm}
\caption{The attack steps of four agents in \textit{env}-1.}
\vspace{-4mm}
\label{attack-steps-chain}
\end{figure}

The training efficiency in \textit{env}-1 can also be observed on the left side of Figure~\ref{attack-steps-chain}, where the two DQRM agents can capture the flag through fewer steps in most episodes compared to the DQN agents. Specifically, the average number of steps taken during the training phase for DQRM-RM1 and DQRM-RM2 is 186.13 and 104.48, respectively, which is approximately half the number of steps taken by the two DQN agents (374.71 and 218.39, respectively).


In Figure~\ref{accumulated-rewards-evaluation-chain}, the x-axis is the logarithmic scale of the step. The trained PT policies of DQRM-RM1, DQRM-RM2, and DQN-RM1 in \textit{env}-1 can achieve higher accumulative rewards in around 50 steps compared to DQN-RM2. DQRM-RM1 and DQRM-RM2 exhibit similar levels of performance, both of which outperform DQN-RM1. On the right side of Figure~\ref{attack-steps-chain}, it can also be observed that DQRM-RM1, DQRM-RM2, and DQN-RM1 can capture the flag using a limited number of steps, which means that their policies are well trained. However, the number of steps in DQN-RM2 has a great variance. Finally, the average number of steps for DQRM-RM1, DQRM-RM2, and DQN-RM1 is 23.48, 21.32, and 29.76, respectively. The average number of steps for DQN-RM2 is 767.46, which reflects poor PT performance.

In \textit{env}-2, DQRM agents also show improved training efficiency and evaluation performance according to Figure~\ref{attack-steps-ctf}. On the left side of Figure~\ref{attack-steps-ctf}, the median values of the steps of DQRM-RM1 (225) and DQRM-RM2 (235) outperform DQN-RM1 (310) and DQN-RM2 (276), respectively. On the right side of Figure~\ref{attack-steps-ctf}, the median value of the steps of DQRM-RM2 (211) is shorter than DQN-RM2 (305), which means it has better evaluation performance. Similarly, DQRM-RM1 (206) also has better evaluation performance than DQN-RM1 (336).

From the previous analysis, we can answer \textbf{RQ1} that RMs can help the agent learn PT policies faster in different environments. 

\subsubsection{Comparison between different RMs} To answer \textbf{RQ2}, we compare the PT performance between DQRM-RM1 and DQRM-RM2 in \textit{env}-1 and \textit{env}-2, respectively. 

From Figure~\ref{accumulated-rewards-training-chain}, DQRM-RM2 demonstrates better training efficiency compared to DQRM-RM1 in \textit{env}-1. We can see that after approximately 80 steps, DQRM-RM2 achieves an average accumulated reward of approximately 15, whereas DQRM-RM1 achieves around 10.

From the left side of Figure~\ref{attack-steps-chain}, the DQRM-RM2 agent requires fewer steps to capture two flags in most episodes compared to the DQRM-RM1 agent in \textit{env}-1. Specifically, the average number of steps taken during the training phase for DQRM-RM1 is 186.13, which is 78\% higher than DQRM-RM2.

From the right side of Figure~\ref{attack-steps-chain}, the average number of attack steps used by DQRM-RM1 and DQRM-RM2 are 23.48 and 21.32, respectively, indicating that DQRM-RM2's PT performance is better than DQRM-RM1 in \textit{env}-1.

\begin{figure}
\centering
\includegraphics[width=0.7\columnwidth]{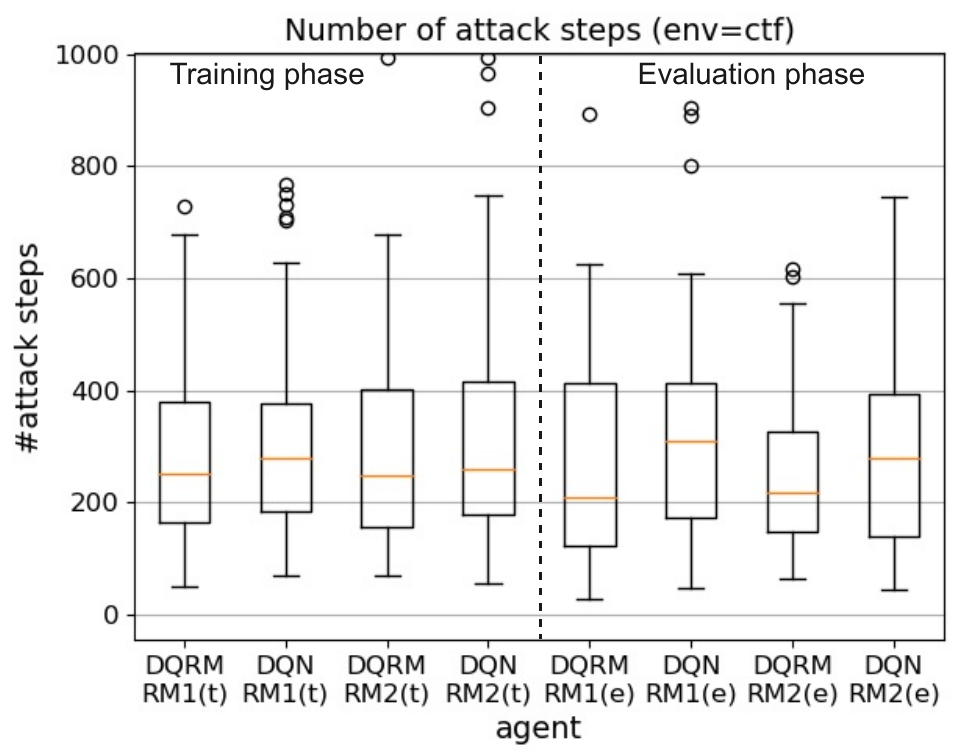}
\vspace{-2mm}
\caption{The attack steps of four agents in \textit{env}-2.}
\vspace{-5mm}
\label{attack-steps-ctf}
\end{figure}

From the left side of Figure~\ref{attack-steps-ctf}, DQRM-RM2 shows fewer average steps (290.85) compared to DQRM-RM1 (302.01) in \textit{env}-2, which means DQRM-RM2 has better training efficiency than DQRM-RM1. On the right side of Figure~\ref{attack-steps-ctf}, DQRM-RM2 has better evaluation performance since DQRM-RM2 takes an average of 295.48 steps, which is fewer than DQRM-RM1 (329.34).

Based on the previous analysis, we can conclude that the PT performance of agents guided by $\mathcal{R}_2$ is more effective than $\mathcal{R}_1$ in different environments since it involves an additional subtask, i.e.,  discovering new nodes, which is more detailed than $\mathcal{R}_1$.

\section{Conclusion and Future Works}\label{conclusion}
In this work, we proposed a knowledge-informed AutoPT framework called DRLRM-PT. This framework utilizes RMs to embed domain knowledge from the field of cybersecurity, which serves as guidelines for training PT policies. We took lateral movement as a case study of PT, and we formulated it as a POMDP guided by RMs. We also designed two RMs based on the MITRE ATT\&CK knowledge base. To train the agent and derive the PT policy, we adopted the DQRM algorithm. The effectiveness of our solution was evaluated using the \textit{CyberBattleSim} platform, from which the experimental results demonstrate that the DQRM agent exhibits a higher training efficiency in PT compared to agents without knowledge embedding. Furthermore, RMs that incorporate more detailed domain knowledge exhibit superior PT performance compared to RMs with simpler knowledge. 

In the future, we will continue investigating MITRE ATT\&CK and Cyber Kill Chain knowledge base to design more sophisticated RMs for PT. The objective is to improve the quality and adaptability of embedded knowledge in various PT scenarios. Additionally, we have plans to expand the scope of PT from lateral movement to more PT applications.



\bibliographystyle{unsrt}
\bibliography{reference}
\vspace{-5mm}

\end{document}